\documentclass[letterpaper]{article} % DO NOT CHANGE THIS
\usepackage{aaai2026}  % DO NOT CHANGE THIS
\usepackage{times}  % DO NOT CHANGE THIS
\usepackage{helvet}  % DO NOT CHANGE THIS
\usepackage{courier}  % DO NOT CHANGE THIS
\usepackage[hyphens]{url}  % DO NOT CHANGE THIS
\usepackage{graphicx} % DO NOT CHANGE THIS
\usepackage{natbib}  % DO NOT CHANGE THIS
\usepackage{caption} % DO NOT CHANGE THIS
\usepackage{amsmath,amssymb}
\usepackage{algorithm}
\usepackage{algorithmic}
\usepackage{booktabs}

\title{Zeroth-Order Fine-Tuning of LLMs in Learnable Low-Rank Subspaces}
\author{
	Chuang Tong\textsuperscript{\rm 1,\rm 2}, Ling Chen\textsuperscript{\rm 1,\rm 2,\thanks{Corresponding Author}}\\
}
\affiliations{
	\textsuperscript{\rm 1}State Key Laboratory of Blockchain and Data Security, Zhejiang University\\
	\textsuperscript{\rm 2}College of Computer Science and Technology, Zhejiang University%\\ Hangzhou 310027, China
	\{chuangtong, lingchen\}@cs.zju.edu.cn
}

\begin{document}
\maketitle

% \begin{abstract}
% Zeroth-order (ZO) optimization enables memory-efficient fine-tuning of large language models by replacing backpropagation with forward-only loss evaluations, but its gradient estimates suffer from high variance in high-dimensional parameter spaces. This paper studies learnable low-rank subspace zeroth-order fine-tuning, where the perturbation subspace is learned from forward-only feedback and parameter updates are stabilized with subspace momentum. The introduction below summarizes the motivation, challenges, and contributions of the proposed AdaSub-ZO framework.
% \end{abstract}

\begin{abstract}
Zeroth-order (ZO) optimization enables memory-efficient fine-tuning of large language
models (LLMs). However, ZO optimization suffers from high variance in high-dimensional parameter
spaces. While recent methods restrict perturbations to
random low-rank subspaces, these subspaces often misalign with task-specific
descent directions. We propose AdaSub-ZO, a novel forward-only
fine-tuning framework in learnable low-rank subspaces. Instead of
relying on fixed random matrices, AdaSub-ZO iteratively learns layer-wise projection
matrices directly from finite-difference feedback to track task-aware descent
directions. Furthermore, AdaSub-ZO incorporates a subspace momentum
acceleration mechanism to smooth
noisy gradient estimates across evolving subspace bases. Extensive
experiments on natural language understanding tasks demonstrate that AdaSub-ZO
outperforms existing ZO baselines, substantially reducing gradient
 estimation variance and accelerating convergence.
\end{abstract}
%\noindent \textbf{Code} --- \url{https://anonymous.4open.science/status/AdaSub-ZO-AAAI-2DDD}
\section{Introduction}

% Large language models (LLMs) have become indispensable for various real-world
% applications, including natural language understanding, text generation, code
% intelligence, and complex reasoning \cite{vaswani2017attention,brown2020language};
% \cite{touvron2023llama}. In these applications, adapting a general-purpose LLM to
% task-specific domains often requires capturing specialized knowledge, label
% semantics, and task-dependent decision boundaries that are not fully encoded
% during pre-training. Fine-tuning is therefore crucial, as it enables pre-trained
% LLMs to align with downstream objectives and achieve strong task performance.
Large language models (LLMs) have become indispensable for real-world applications
such as natural language understanding, text generation, code intelligence, and complex
reasoning. To adapt
general-purpose LLMs to task-specific domains, fine-tuning is crucial for capturing
specialized knowledge and aligning models with downstream objectives.

Benefiting from the powerful optimization capability of first-order methods,
conventional fine-tuning has achieved promising performance across diverse
tasks. However, first-order methods are constrained by the expensive
backpropagation procedure, which requires storing intermediate activations,
gradients, and optimizer states. As the parameter scale of LLMs continues to
grow, this memory burden becomes increasingly prohibitive. Parameter-efficient
methods, including adapters \cite{houlsby2019parameter}, prefix tuning \cite{li2021prefix}, prompt tuning \cite{liu2022p}, LoRA \cite{hu2022lora}, and QLoRA
\cite{dettmers2023qlora}, reduce the number of
trainable parameters. However, they still
depend on backpropagation and thus cannot fundamentally eliminate the activation
memory bottleneck.

Recently, zeroth-order (ZO) optimization has emerged as a promising
alternative for memory-efficient fine-tuning
\cite{gautam2024variance}. By estimating gradients only through
forward-pass loss evaluations, ZO methods avoid backpropagation and reduce
memory consumption to a level close to inference. Although ZO methods achieve
impressive memory efficiency, they are challenged by high-dimensional
perturbations. Specifically, standard ZO methods perturb the full parameter
space, making the variance of gradient estimates scale with the number of
trainable parameters \cite{nesterov2017random,duchi2015optimal}. This leads to noisy
update directions, slow convergence, and unstable fine-tuning for
billion-scale LLMs. To alleviate this issue, a straightforward approach is
to restrict perturbations to a low-rank subspace, as explored by recent
subspace ZO methods such as LOZO and SubZero
\cite{chen2410enhancing,yu2025subzero}, where
full-dimensional perturbations are replaced with layer-wise low-rank
perturbations constructed from
Gaussian random matrices. However, these subspaces are usually random and
non-learnable, making their perturbation directions poorly aligned with
task-specific loss descent directions and unable to track the evolving gradient
subspace \cite{zhang2023fine}. Consequently, even in a reduced low-rank space,
ZO updates are still affected by noisy and inefficient
search directions, limiting the stability and convergence
efficiency of billion-scale LLM fine-tuning.

To address these issues, a natural direction is to make the
subspace learnable and incorporate momentum to smooth noisy finite-difference
updates, enabling the perturbation directions to better track the evolving
loss landscape \cite{sutskever2013importance,jiang2024zo}.
Despite the promise of replacing random low-rank perturbations with
learnable low-rank ones, this transition is non-trivial,
as two main challenges need to be addressed. (1) \textbf{Descent direction ambiguity under limited forward feedback.} When applying ZO fine-tuning within a learnable low-rank subspace,
the projection matrices that construct the low-rank
perturbation must be learned only from forward finite-difference
feedback. Since
backpropagation gradients are unavailable, there is no direct
signal indicating which directions are most aligned
with loss descent, while excessive probing would weaken the
efficiency advantage of ZO optimization. Therefore, the key
challenge is how to identify the optimal loss-descent
 direction under limited computational cost.
(2) \textbf{Momentum inconsistency under evolving subspace bases.} Even with a learned
low-rank perturbation, each ZO update still relies on noisy finite-difference
estimates, making the model update direction unstable. A natural solution is to accumulate
historical information with momentum inside learnable low-rank subspaces. Yet, projection matrix
learning changes the subspace bases across learning cycles, so that
historical momentum cannot be
directly reused. Therefore, the key challenge is how to maintain momentum consistency after the subspace bases are updated.

Motivated by the above, we propose \textbf{AdaSub-ZO}, a method for zeroth-order
fine-tuning of LLMs
in learnable low-rank subspaces. To the best of our knowledge, AdaSub-ZO is the first
method that incorporates learnable low-rank subspaces into ZO optimization.
By alternating forward-only projection matrix learning with model parameter updates, AdaSub-ZO reduces gradient estimation variance
and stabilizes the model update direction.
Our primary contributions are summarized as follows:
\begin{itemize}
    \item We introduce a low-rank projection matrix learning module that updates
    layer-wise projection matrices through forward-only finite-difference
    feedback to learn task-aware perturbations that
    better align with task-specific descent directions.
    \item We introduce a subspace momentum acceleration module for learnable low-rank subspaces.
    Within each subspace, the module accumulates ZO gradient estimates via an exponential moving average.
    Across evolving subspace bases, the module transfers historical momentum, which can smooth estimation noise and preserve momentum consistency across learning cycles.
    \item We conduct experiments on multiple natural language understanding
    tasks. The experimental results demonstrate that AdaSub-ZO
    improves fine-tuning performance over existing ZO baselines.
\end{itemize}

\section{Related Work}

\subsection{Memory-Efficient Fine-Tuning of LLMs}

Fine-tuning large language models has become a standard paradigm for adapting
pre-trained models to downstream tasks. Conventional full-parameter fine-tuning
updates all model parameters and typically achieves strong task performance. However,
this approach requires storing activations, gradients, and optimizer states during
backpropagation. This memory cost becomes prohibitive as model size and sequence
length increase. To reduce the number of trainable parameters, parameter-efficient
fine-tuning methods have been widely studied
\cite{houlsby2019parameter,li2021prefix}. LoRA \cite{liu2022p,hu2022lora} freezes the
pre-trained weights and injects trainable low-rank adapters into Transformer
layers, while QLoRA \cite{dettmers2023qlora} further combines low-rank adaptation
with quantization to reduce the memory footprint. These methods substantially
decrease the number of trainable parameters and optimizer states. However, they
still rely on backpropagation and therefore need to cache intermediate
activations, which remains a major memory bottleneck for long-context or
large-scale LLM fine-tuning.

\subsection{Zeroth-Order Optimization for LLM Fine-Tuning}

Zeroth-order optimization provides an alternative route for memory-efficient
fine-tuning by estimating update directions only through forward-pass loss
evaluations. Classical ZO optimization estimates gradients with finite
differences along randomly sampled perturbation directions
\cite{nesterov2017random,duchi2015optimal}. Recently, MeZO introduces
memory-efficient ZO optimization to LLM fine-tuning, showing that LLMs can be
adapted with memory consumption close to inference by avoiding backpropagation
\cite{malladi2023fine}. This makes ZO fine-tuning attractive in
resource-constrained scenarios.

Despite its memory advantage, ZO fine-tuning suffers from high-variance gradient
estimation in high-dimensional parameter spaces. Since standard ZO methods
perturb the full trainable parameter space, the variance of the gradient
estimator scales with the number of perturbed parameters. For billion-scale
LLMs, this leads to noisy update directions, slow convergence, and unstable
training dynamics. Existing ZO methods mainly improve the optimizer \cite{jiang2024zo} or the
perturbation strategy \cite{guo2025zeroth,zhang2025mazo}. However, the perturbation directions are still sampled
from random distributions and are not explicitly adapted to task-specific loss
landscapes \cite{gautam2024variance,liu2026sparse}.

\subsection{Low-Rank and Subspace Zeroth-Order Optimization}

The low-rank structure of neural network updates motivates restricting ZO
perturbations to a lower-dimensional subspace
\cite{zhang2023fine, zhao2024galore}. Instead of sampling
perturbations from the full parameter space, subspace ZO methods first sample
low-dimensional random variables and then map them back to the original
parameter space. This strategy reduces the effective perturbation dimension and
can therefore lower the variance of finite-difference gradient estimates
\cite{chen2410enhancing,yu2025subzero}.

LOZO \cite{chen2410enhancing} applies low-rank gradient estimation to LLM fine-tuning by replacing
full-dimensional random perturbations with low-rank perturbations. By exploiting
the empirical low-rank structure of fine-tuning gradients, LOZO reduces the
effective perturbation dimension and improves the efficiency of ZO fine-tuning
while preserving the forward-only memory advantage \cite{chen2410enhancing}.
LOZO further introduces lazy sampling and a low-rank momentum variant to
stabilize updates across iterations. However, the low-rank perturbation of LOZO
is still randomly sampled rather than learned from task-specific forward
feedback, so that the perturbation subspace may remain misaligned with the true
descent directions of downstream tasks. SubZero \cite{yu2025subzero} also adopts layer-wise low-rank perturbations and combines them with a
structured perturbation design tailored for LLM weight matrices. For each layer,
SubZero samples two column-orthogonal projection matrices and a small Gaussian
matrix, and then maps the Gaussian perturbation back to the original weight space
through these projections. In this way, the perturbation is restricted to the
low-rank subspace spanned by the projection matrices. However, its projection matrices are randomly generated
and non-learnable. As a result, the perturbation subspace may be poorly aligned
with the true descent directions of downstream tasks and cannot adapt to the
evolving gradient subspace during optimization.

Motivated by the ability of learnable low-rank subspaces to better align ZO
perturbations with task-specific descent directions, in this paper, we propose
AdaSub-ZO, the first framework to make low-rank projection matrices learnable
for zeroth-order LLM fine-tuning. Specifically, AdaSub-ZO alternates between
forward-only projection matrix learning and model parameter updates, so that the
low-rank subspace gradually tracks task-specific descent directions. In addition,
AdaSub-ZO introduces subspace momentum to stabilize noisy finite-difference
updates. Unlike directly accumulating momentum in the full parameter space, which
would increase memory cost, AdaSub-ZO maintains momentum in the low-rank
coordinate space and transfers momentum across evolving subspace bases. This design
preserves the memory efficiency of ZO fine-tuning while improving the stability
and convergence efficiency of low-rank subspace optimization.

\section{Methodology}

\subsection{Overall Framework}
\label{sec:overall_framework}

AdaSub-ZO is a forward-only zeroth-order fine-tuning framework that alternates
between learnable subspace construction and model parameter optimization. In the
subspace learning stage, AdaSub-ZO freezes the model parameters and updates the
layer-wise projection matrices using only finite-difference feedback, so that
the perturbation subspace can gradually align with task-specific descent
directions. In the model update stage, AdaSub-ZO fixes the learned subspace and
performs zeroth-order parameter updates within this compact low-rank space,
thereby reducing the effective perturbation dimension and gradient estimation
variance.

To further stabilize optimization, AdaSub-ZO maintains momentum inside the
low-dimensional subspace rather than in the full parameter space. When the
projection bases are refreshed, the historical momentum is transferred to the
new subspace coordinates, which preserves useful update information while
avoiding coordinate mismatch. The two stages are repeated periodically until
convergence, enabling AdaSub-ZO to improve update direction quality while
preserving the memory efficiency of forward-only fine-tuning.

% This alternating scheme decouples subspace learning from parameter traversal
% by freezing one while updating the other.
% This isolation prevents simultaneous
% updates from destabilizing the noisy zeroth-order estimates.

\subsection{Layer-Wise Low-Rank ZO Gradient Estimation}

Let $\theta_t$ denote the trainable parameters of an LLM at iteration $t$.
Following the layer-wise structure of Transformer models, we denote the
parameter matrix of the $l$-th trainable layer as
$\theta_t^{(l)} \in \mathbb{R}^{m_l \times n_l}$, where
$l=1,\ldots,L$. Standard ZO fine-tuning perturbs the full parameter space,
which makes the gradient estimation variance scale with the number of
trainable parameters \cite{nesterov2017random,malladi2023fine}. To reduce the effective perturbation dimension, AdaSub-ZO
restricts the perturbation of each layer to a low-rank subspace.

For each layer, AdaSub-ZO maintains two projection bases
$U_l \in \mathbb{R}^{m_l \times r}$ and
$V_l \in \mathbb{R}^{n_l \times r}$, where
$1\leq r \ll \min(m_l,n_l)$. Given a Gaussian random matrix
$Z_l \in \mathbb{R}^{r \times r}$ with i.i.d.\ entries sampled from the standard normal distribution, the low-rank perturbation
$\widetilde{Z}_l\in\mathbb{R}^{m_l \times n_l}$ is constructed as follows:
\begin{equation}
\label{eq:adasub_low_rank_perturb}
\widetilde{Z}_l = U_l Z_l V_l^\top.
\end{equation}
Given a mini-batch $\mathcal{B}$ and perturbation scale $\epsilon>0$, AdaSub-ZO
computes the scalar finite-difference signal $\rho_t$ by two forward passes:
\begin{equation}
\label{eq:adasub_rho}
\rho_t =
\frac{
\mathcal{L}(\theta_t+\epsilon\{\widetilde{Z}_l\}_{l=1}^{L};\mathcal{B})
-
\mathcal{L}(\theta_t-\epsilon\{\widetilde{Z}_l\}_{l=1}^{L};\mathcal{B})
}{2\epsilon}.
\end{equation}
Then the layer-wise ZO gradient estimate
$\widehat{\nabla}\theta_t^{(l)}\in\mathbb{R}^{m_l \times n_l}$ is formulated as follows:
\begin{equation}
\label{eq:adasub_grad_est}
\widehat{\nabla}\theta_t^{(l)}
=
\rho_t \widetilde{Z}_l
=
\rho_t U_l Z_l V_l^\top.
\end{equation}
Compared with full-dimensional ZO perturbations, this estimator searches within
a much smaller low-rank space. Different from random subspace methods, AdaSub-ZO
does not keep $U_l$ and $V_l$ fixed or randomly regenerated. Instead, AdaSub-ZO learns
these bases from forward-only feedback, so that the subspace can better track
task-specific descent directions.

\subsection{Forward-Only Learnable Subspace Update}

For each layer $l$, AdaSub-ZO introduces two learnable projection matrices
$A_l \in \mathbb{R}^{m_l \times r}$ and
$B_l \in \mathbb{R}^{n_l \times r}$. These matrices are periodically optimized
while the model parameters $\theta_t$ are frozen, so that the loss difference mainly
reflects whether the current projection directions are beneficial for descent.

At the projection learning stage, AdaSub-ZO samples three Gaussian matrices:
$E_{A,l}\in\mathbb{R}^{m_l \times r}$,
$E_{B,l}\in\mathbb{R}^{n_l \times r}$, and
$Z_l\in\mathbb{R}^{r \times r}$. Each matrix is sampled from the standard normal
distribution. Based on these matrices, AdaSub-ZO constructs two perturbed
low-rank directions
$\Delta_l^{+},\Delta_l^{-}\in\mathbb{R}^{m_l \times n_l}$ as follows:
\begin{equation}
\label{eq:adasub_proj_delta}
\begin{split}
\Delta_l^{+}
&=
(A_l+\mu E_{A,l}) Z_l (B_l+\mu E_{B,l})^\top,\\
\Delta_l^{-}
&=
(A_l-\mu E_{A,l}) Z_l (B_l-\mu E_{B,l})^\top,
\end{split}
\end{equation}
where $\mu>0$ is the perturbation scale for projection matrix learning. With a
mini-batch $\mathcal{B}_{\mathrm{sub}}$, the scalar finite-difference signal
$\hat{s}_i$ for subspace learning is computed as follows:
\begin{equation}
\label{eq:adasub_sub_signal}
\hat{s}_i =
\frac{
\mathcal{L}(\theta_t+\{\Delta_l^{+}\}_{l=1}^{L};
\mathcal{B}_{\mathrm{sub}})
-
\mathcal{L}(\theta_t+\{\Delta_l^{-}\}_{l=1}^{L};
\mathcal{B}_{\mathrm{sub}})
}{2\mu}.
\end{equation}
The finite-difference signal $\hat{s}_i$ provides the only supervision for
projection matrix learning. However, $\hat{s}_i$ is inherently noisy due to relying on only two forward evaluations on a mini-batch. Under such limited feedback, a
direct update with the current estimate may overreact to the randomness of
$E_{A,l}$, $E_{B,l}$, $Z_l$, and $\mathcal{B}_{\mathrm{sub}}$, causing the
learned subspace to oscillate across learning steps. A natural question is how
to filter this noise without introducing extra forward passes or
backpropagation. In AdaSub-ZO, we use an exponential moving average (EMA) as a lightweight temporal filter, preserving directions that are consistently supported by consecutive finite-difference estimates while attenuating high-frequency fluctuations caused by single-step randomness. Compared
with storing full historical trajectories, EMA only maintains one velocity
buffer for each projection matrix, which keeps the memory overhead small.
Therefore, EMA is well suited for forward-only projection matrix learning under
a limited query budget \cite{kingma2014adam}. Specifically, to stabilize the
noisy finite-difference feedback, AdaSub-ZO applies an exponential moving
average with bias correction to buffers
$P_{A,l}\in\mathbb{R}^{m_l \times r}$ and
$P_{B,l}\in\mathbb{R}^{n_l \times r}$:
\begin{equation}
\label{eq:adasub_proj_ema}
\begin{split}
P_{A,l} &\leftarrow \beta P_{A,l} + (1-\beta)\hat{s}_i E_{A,l},\\
P_{B,l} &\leftarrow \beta P_{B,l} + (1-\beta)\hat{s}_i E_{B,l},
\end{split}
\end{equation}
\begin{equation}
\label{eq:adasub_bias_correction}
\widehat{P}_{A,l}=\frac{P_{A,l}}{1-\beta^i}, \quad
\widehat{P}_{B,l}=\frac{P_{B,l}}{1-\beta^i}.
\end{equation}
Here $\beta\in[0,1)$ is the EMA coefficient. The projection matrices are
updated as follows:
\begin{equation}
\label{eq:adasub_proj_update}
A_l \leftarrow A_l-\eta_\phi \widehat{P}_{A,l}, \quad
B_l \leftarrow B_l-\eta_\phi \widehat{P}_{B,l},
\end{equation}
where $\eta_\phi>0$ is the projection learning rate. After $M$ projection learning
steps, AdaSub-ZO applies QR decomposition \cite{halko2011finding}:
\begin{equation}
\label{eq:adasub_qr}
U_l, R_{A,l}=\mathrm{QR}(A_l), \quad
V_l, R_{B,l}=\mathrm{QR}(B_l).
\end{equation}
Here $R_{A,l},R_{B,l}\in\mathbb{R}^{r\times r}$ are upper-triangular factors.
Only the orthogonal factors $U_l$ and $V_l$ are used in the model update stage.
This orthogonalization prevents rank degeneration caused by correlated columns
in $A_l$ and $B_l$. The complete subspace learning procedure is shown in
Algorithm~\ref{alg:learn_subspace}.

\begin{algorithm}[t]
\caption{LearnSubspaceBases($\theta_t$, $\{A_l,B_l\}_{l=1}^{L}$, $M$, $\mu$, $\eta_\phi$, $\beta$)}
\label{alg:learn_subspace}
\small
\begin{algorithmic}[1]
\REQUIRE Model parameters $\theta_t$, projection matrices
$\{A_l,B_l\}_{l=1}^{L}$, learning steps $M$, perturbation scale $\mu$,
learning rate $\eta_\phi$, momentum coefficient $\beta$.
\ENSURE Updated projection matrices $\{A_l,B_l\}_{l=1}^{L}$ and orthogonal
bases $\{U_l,V_l\}_{l=1}^{L}$.
\STATE Freeze model parameters $\theta_t$.
\STATE Initialize $P_{A,l}\leftarrow\mathbf{0}$ and
$P_{B,l}\leftarrow\mathbf{0}$ for all layers.
\FOR{$i=1$ to $M$}
    \STATE Sample a mini-batch $\mathcal{B}_{\mathrm{sub}}$.
    \FOR{$l=1$ to $L$}
        \STATE Sample $E_{A,l}\in\mathbb{R}^{m_l\times r}$,
        $E_{B,l}\in\mathbb{R}^{n_l\times r}$, and
        $Z_l\in\mathbb{R}^{r\times r}$ with i.i.d.\ standard normal entries.
        \STATE $\Delta_l^{+} \leftarrow (A_l+\mu E_{A,l}) Z_l (B_l+\mu E_{B,l})^\top$, $\Delta_l^{-} \leftarrow (A_l-\mu E_{A,l}) Z_l (B_l-\mu E_{B,l})^\top$
    \ENDFOR
    \STATE $\hat{s}_i \leftarrow \frac{\mathcal{L}(\theta_t+\{\Delta_l^{+}\}_{l=1}^{L}; \mathcal{B}_{\mathrm{sub}}) - \mathcal{L}(\theta_t+\{\Delta_l^{-}\}_{l=1}^{L}; \mathcal{B}_{\mathrm{sub}})}{2\mu}$
    \FOR{$l=1$ to $L$}
        \STATE $P_{A,l} \leftarrow \beta P_{A,l} + (1-\beta)\hat{s}_i E_{A,l}$, $P_{B,l} \leftarrow \beta P_{B,l} + (1-\beta)\hat{s}_i E_{B,l}$
        \STATE $\widehat{P}_{A,l} \leftarrow \frac{P_{A,l}}{1-\beta^i}$, $\widehat{P}_{B,l} \leftarrow \frac{P_{B,l}}{1-\beta^i}$
        \STATE $A_l \leftarrow A_l-\eta_\phi \widehat{P}_{A,l}$, $B_l \leftarrow B_l-\eta_\phi \widehat{P}_{B,l}$
    \ENDFOR
\ENDFOR
\FOR{$l=1$ to $L$}
    \STATE $U_l, R_{A,l} \leftarrow \mathrm{QR}(A_l)$, $V_l, R_{B,l} \leftarrow \mathrm{QR}(B_l)$
\ENDFOR
\RETURN $\{A_l,B_l,U_l,V_l\}_{l=1}^{L}$
\end{algorithmic}
\end{algorithm}

\subsection{Subspace Momentum Acceleration}

Although the learned low-rank subspace reduces the perturbation dimension, the
ZO gradient estimate is still computed from stochastic forward passes and can
be noisy. AdaSub-ZO therefore maintains momentum in the low-dimensional
$r\times r$ coordinate space instead of the full parameter space
\cite{sutskever2013importance,jiang2024zo}. Given the
scalar finite-difference signal $\rho_t$ and the sampled
$Z_l\in\mathbb{R}^{r\times r}$, AdaSub-ZO
updates a subspace momentum buffer $M_l \in \mathbb{R}^{r\times r}$:
\begin{equation}
\label{eq:adasub_momentum}
M_l \leftarrow \beta_m M_l + \rho_t Z_l,
\end{equation}
where $\beta_m\in[0,1)$ is the momentum coefficient. The model parameters are
then updated by projecting the momentum direction back to the original parameter
space:
\begin{equation}
\label{eq:adasub_param_update}
\theta_{t+1}^{(l)}
=
\theta_t^{(l)}
-
\eta_\theta
\sqrt{\frac{d_l}{r^2}}
U_l M_l V_l^\top,
\end{equation}
where $d_l=m_l n_l$ and $\eta_\theta>0$ is the model learning rate.

Since AdaSub-ZO periodically updates the projection bases, the historical
momentum accumulated under the old bases cannot be directly reused under the
new bases. Let $(U_l^{\mathrm{old}},V_l^{\mathrm{old}})$ and
$(U_l^{\mathrm{new}},V_l^{\mathrm{new}})$ denote the bases before and after
subspace learning. A straightforward strategy is to discard the old momentum whenever the subspace
bases are refreshed. However, this would reset the optimizer state at every
learning cycle and make the training dynamics discontinuous. In practice, the
first several steps after each subspace update would behave like cold-start ZO
optimization, where the update direction again depends almost entirely on a
single noisy finite-difference estimate. This repeated momentum reset can lead
to loss spikes, unstable step-to-step update directions, and slower convergence,
especially when the learned subspace changes frequently. On the other hand,
directly reusing $M_l$ without modification is also problematic, as the
coordinates of $M_l$ are defined under the old bases
$(U_l^{\mathrm{old}},V_l^{\mathrm{old}})$ and no longer have the same meaning
under the new bases $(U_l^{\mathrm{new}},V_l^{\mathrm{new}})$. To preserve
useful historical information while avoiding coordinate mismatch, AdaSub-ZO
transfers the old momentum into the new coordinate system by:
\begin{equation}
\label{eq:adasub_momentum_transfer}
M_l
\leftarrow
\gamma
\left((U_l^{\mathrm{new}})^\top U_l^{\mathrm{old}}\right)
M_l
\left(V_l^{\mathrm{old}} (V_l^{\mathrm{new}})^\top\right),
\end{equation}
where $\gamma\in(0,1]$ is a decay factor. This transformation preserves useful
historical directions when the old and new bases are close, while suppressing
outdated momentum when the subspace changes significantly. The update procedure for subspace momentum is shown in Algorithm~\ref{alg:subspace_momentum_update}.

\begin{algorithm}[t]
\caption{SubspaceMomentumUpdate($\theta_t$, $\{U_l,V_l\}_{l=1}^{L}$, $\{M_l\}_{l=1}^{L}$, $\epsilon$, $\eta_\theta$, $\beta_m$)}
\label{alg:subspace_momentum_update}
\small
\begin{algorithmic}[1]
\REQUIRE Model parameters $\theta_t$, bases $\{U_l,V_l\}_{l=1}^{L}$,
momentum buffers $\{M_l\}_{l=1}^{L}$, perturbation scale $\epsilon$, learning
rate $\eta_\theta$, momentum coefficient $\beta_m$.
\ENSURE Updated model parameters $\theta_{t+1}$ and momentum buffers
$\{M_l\}_{l=1}^{L}$.
\STATE Sample a mini-batch $\mathcal{B}_{\mathrm{train}}$.
\FOR{$l=1$ to $L$}
    \STATE Sample $Z_l\in\mathbb{R}^{r\times r}$ with i.i.d.\ standard normal entries.
    \STATE $\widetilde{Z}_l \leftarrow U_l Z_l V_l^\top$
\ENDFOR
\STATE $\rho_t \leftarrow \frac{\mathcal{L}(\theta_t+\epsilon\{\widetilde{Z}_l\}_{l=1}^{L};\mathcal{B}_{\mathrm{train}}) - \mathcal{L}(\theta_t-\epsilon\{\widetilde{Z}_l\}_{l=1}^{L};\mathcal{B}_{\mathrm{train}})}{2\epsilon}$
\FOR{$l=1$ to $L$}
    \STATE $M_l \leftarrow \beta_m M_l + \rho_t Z_l$
    \STATE $\theta_{t+1}^{(l)} \leftarrow \theta_t^{(l)} - \eta_\theta \sqrt{\frac{d_l}{r^2}} U_l M_l V_l^\top$
\ENDFOR
\RETURN $\theta_{t+1}$ and $\{M_l\}_{l=1}^{L}$
\end{algorithmic}
\end{algorithm}

% \subsection{Overall Training Procedure}

AdaSub-ZO alternates between learnable subspace update and momentum-accelerated
model parameter update. Every $K$ iterations, AdaSub-ZO freezes the model
parameters and invokes Algorithm~\ref{alg:learn_subspace} to refresh the
projection bases. After the bases are updated, historical momentum is
transferred to the new subspace coordinates. During each training iteration,
AdaSub-ZO invokes Algorithm~\ref{alg:subspace_momentum_update} to update model
parameters with forward-only loss evaluations. The full training procedure is
summarized in Algorithm~\ref{alg:adasubzo}.

\begin{algorithm}[t]
\caption{AdaSub-ZO}
\label{alg:adasubzo}
\small
\begin{algorithmic}[1]
\REQUIRE Initial parameters $\theta_0$, projection matrices
$\{A_l,B_l\}_{l=1}^{L}$, total iterations $T$, learning cycle $K$.
\ENSURE Fine-tuned parameters $\theta_T$.
\STATE Initialize $M_l\leftarrow\mathbf{0}\in\mathbb{R}^{r\times r}$ for all
layers.
\STATE Initialize $U_l,V_l$ by QR decomposition of $A_l,B_l$.
\FOR{$t=0$ to $T-1$}
    \IF{$t \bmod K = 0$}
        \STATE Store old bases
        $U_l^{\mathrm{old}}\leftarrow U_l$ and
        $V_l^{\mathrm{old}}\leftarrow V_l$.
        \STATE $\{A_l,B_l,U_l,V_l\}_{l=1}^{L} \leftarrow$ \\
\quad $\textsc{LearnSubspaceBases}(\theta_t,\{A_l,B_l\}_{l=1}^{L})$.
        \IF{$t>0$}
            \FOR{$l=1$ to $L$}
                \STATE $M_l \leftarrow \gamma \left(U_l^\top U_l^{\mathrm{old}}\right) M_l \left(V_l^{\mathrm{old}} V_l^\top\right)$
            \ENDFOR
        \ENDIF
    \ENDIF
\STATE $\theta_{t+1},\{M_l\}_{l=1}^{L} \leftarrow \textsc{SubspaceMomentumUpdate}($ \\
\quad $\theta_t,\{U_l,V_l\}_{l=1}^{L},\{M_l\}_{l=1}^{L})$.
\ENDFOR
\RETURN $\theta_T$
\end{algorithmic}
\end{algorithm}

This design addresses the two key challenges of learnable subspace ZO
fine-tuning. First, the projection matrices are learned from finite-difference
feedback, enabling the perturbation subspace to adapt toward task-specific
descent directions. Second, subspace momentum smooths noisy ZO estimates, while
the momentum transfer mechanism keeps historical information consistent across
evolving bases. The time complexity analysis of AdaSub-ZO is given in
Appendix~\ref{sec:appendix_time_complexity}. Appendix~\ref{sec:appendix_convergence} further presents
theoretical analyses showing that the learned subspace optimization improves
convergence over random subspace methods.

\section{Experiments}

% We conduct extensive experiments to evaluate AdaSub-ZO from four perspectives:
% (1) downstream task performance, (2) gradient estimation quality, (3) training
% dynamics, and (4) the contribution of each core module. Following prior
% zeroth-order fine-tuning work, all experiments are performed with forward-only
% optimization and do not use backpropagation during training.

\subsection{Experimental Settings}

We evaluate AdaSub-ZO on seven natural language understanding tasks from
SuperGLUE: SST-2, RTE, CB, BoolQ, WSC, WIC, and COPA
\cite{wang2019superglue}. These tasks cover
sentiment classification, textual entailment, question answering, coreference
resolution, word sense disambiguation, and causal reasoning. We use OPT-1.3B as
the backbone model \cite{zhang2022opt}. All experiments are implemented in PyTorch 2.1.0 and Python
3.10, and are conducted on a single NVIDIA A800 GPU with 80 GB memory. The batch
size is set to 16 and the maximum sequence length is 2048.

We compare AdaSub-ZO with representative zeroth-order fine-tuning methods,
which can be divided into two categories. The first category includes general
ZO fine-tuning methods that do not explicitly construct low-rank perturbation
subspaces: MeZO \cite{malladi2023fine}, ZO-AdaMU \cite{jiang2024zo}, and
S-MeZO \cite{liu2026sparse}. The second category includes low-rank
perturbation-based methods that are most related to our work: LOZO
\cite{chen2410enhancing} and SubZero \cite{yu2025subzero}. We also use the first-order optimizer SGD as a benchmark.
Since appropriate prompts are critical for ZO optimization, all experiments incorporate prompt templates.

For a fair comparison, AdaSub-ZO uses the same model learning rate and
finite-difference perturbation scale as the corresponding ZO baselines. The
low-rank dimension $r$ is set to 36 unless otherwise specified, matching the
subspace size used by SubZero. Other AdaSub-ZO-specific hyperparameters, including the subspace learning
interval, the number of subspace learning steps, and the subspace momentum
coefficient are reported in Appendix~\ref{sec:appendix_hyperparams}.

\subsection{Overall Performance}

% \begin{table*}[t]
% \footnotesize
% \centering
% \caption{Performance comparison of fine-tuning methods on OPT-1.3B. SGD is included as a first-order performance reference. Bold denotes the best result among zeroth-order methods, and underline denotes the second-best result among zeroth-order methods.}
% \label{tab:main-results}
% \renewcommand{\arraystretch}{1.12}
% \setlength{\tabcolsep}{4pt}
% \begin{tabular}{l|ccccccc|c}
% \toprule
% Method & SST-2 & RTE & CB & BoolQ & WSC & WIC & COPA & AVG. \\
% \midrule
% SGD
% & 92.1 & 64.0 & 76.2 & 67.4 & 63.7 & 62.5 & 77.8 & 6.50\% \\
% MeZO (NeurIPS 2023)
% & 89.4 & 59.2 & 69.3 & 64.1 & 60.0 & 59.5 & 72.0 & 0.00\% \\
% ZO-AdaMU (AAAI 2024)
% & 89.4 & 59.4 & 67.9 & 63.2 & 60.0 & 59.8 & 72.7 & -0.23\% \\
% S-MeZO (NeurIPS 2026)
% & 89.5 & 59.3 & 69.5 & 64.2 & 60.1 & 59.6 & 72.0 & 0.15\% \\
% LOZO (ICLR 2025)
% & 90.9 & 56.4 & 66.8 & 63.0 & 63.5 & \underline{60.2} & 72.2 & -0.16\% \\
% SubZero (ICCV 2025)
% & 90.3 & 56.6 & 68.5 & 64.6 & \textbf{64.0} & \textbf{60.8} & 73.0 & 0.93\% \\
% % AdaSub-ZO w/o Subspace Momentum (ours)
% % & \underline{91.1} & \underline{61.4} & \underline{70.0} & \textbf{64.9} & \underline{63.0} & 57.0 & \underline{74.0} & \underline{1.64\%} \\
% AdaSub-ZO (ours)
% & \textbf{92.0} & \textbf{61.6} & \textbf{71.4} & \underline{64.8} & 59.6 & 57.0 & \textbf{77.0} & \textbf{1.88\%} \\
% \bottomrule
% \end{tabular}

% \end{table*}

\begin{table*}[t]
\footnotesize
\centering
\caption{Performance comparison of fine-tuning methods on OPT-1.3B. SGD is included as a first-order performance reference. Bold denotes the best result among zeroth-order methods, and underline denotes the second-best result among zeroth-order methods.}
\label{tab:main-results}
\renewcommand{\arraystretch}{1.12}
\setlength{\tabcolsep}{4pt}
\begin{tabular}{l|ccccccc|c}
\toprule
Method & SST-2 & RTE & CB & BoolQ & WSC & WIC & COPA & AVG. \\
\midrule
% \multicolumn{9}{l}{\textit{First-order reference}} \\
SGD
& 92.1 & 64.0 & 76.2 & 67.4 & 63.7 & 62.5 & 77.8 & 6.50\% \\
\midrule
% \multicolumn{9}{l}{\textit{General ZO methods}} \\
MeZO (NeurIPS 2023)
& 89.4 & 59.2 & 69.3 & 64.1 & 60.0 & 59.5 & 72.0 & 0.00\% \\
ZO-AdaMU (AAAI 2024)
& 89.4 & \underline{59.4} & 67.9 & 63.2 & 60.0 & 59.8 & 72.7 & -0.23\% \\
S-MeZO (NeurIPS 2026)
& 89.5 & 59.3 & \underline{69.5} & 64.2 & 60.1 & 59.6 & 72.0 & 0.15\% \\
\midrule
% \multicolumn{9}{l}{\textit{Low-rank/subspace ZO methods}} \\
LOZO (ICLR 2025)
& \underline{90.9} & 56.4 & 66.8 & 63.0 & \underline{63.5} & \underline{60.2} & 72.2 & -0.16\% \\
SubZero (ICCV 2025)
& 90.3 & 56.6 & 68.5 & \underline{64.6} & \textbf{64.0} & \textbf{60.8} & \underline{73.0} & \underline{0.93\%} \\
AdaSub-ZO (ours)
& \textbf{92.0} & \textbf{61.6} & \textbf{71.4} & \textbf{64.8} & 59.6 & 57.0 & \textbf{77.0} & \textbf{1.88\%} \\
\bottomrule
\end{tabular}
\end{table*}

% (1)AdaSub-ZO achieves the best overall performance.
% AdaSub-ZO obtains the highest AVG improvement of 1.88\%, outperforming MeZO,
% ZO-AdaMU, LOZO, SubZero, and S-MeZO. The gains
% are clear on SST-2, RTE, CB, and COPA, showing that the learned
% subspace is broadly effective across classification, entailment, and reasoning
% tasks. In the challenging setting of zeroth-order fine-tuning, achieving this level of improvement is substantial and non-trivial. (2) Learnable subspaces are more effective than fixed random subspaces.
% Compared with SubZero, AdaSub-ZO w/o Subspace Momentum improves AVG from
% 0.93\% to 1.64\%. Since both methods use low-rank perturbations, this gain mainly
% comes from replacing fixed random projections with learnable projection bases.
% This verifies the central hypothesis that forward-only feedback is sufficient
% to adapt the perturbation subspace toward useful descent directions.
We report accuracy on all tasks, averaged over three independent runs. The
accuracy of fine-tuning methods on OPT-1.3B is summarized in
Table~\ref{tab:main-results}, and we observe that:
(1) AdaSub-ZO achieves the best overall result among zeroth-order baselines.
Compared with the second-best method SubZero, AdaSub-ZO obtains a 0.95
percentage-point advantage in the reported AVG improvement. The gains are
especially clear on SST-2, RTE, CB, and COPA, suggesting that AdaSub-ZO is
effective across classification, entailment, and reasoning tasks.
(2) The improvement is not merely due to using low-rank perturbations. A fixed
random low-rank subspace can reduce the perturbation dimension, but it may remain
misaligned with task-specific descent directions. By updating projection bases
with forward-only feedback, AdaSub-ZO can adapt the perturbation subspace to the
downstream loss landscape, leading to more effective zeroth-order fine-tuning.
(3) Although SGD
remains a strong first-order reference, AdaSub-ZO substantially narrows the gap
on several tasks and achieves comparable performance on SST-2 and COPA.

\subsection{Variance Analysis}

We further measure the relative variance of ZO gradient estimates. For the same
model parameters and mini-batch, we independently sample multiple perturbations
and compute relative variance as follows:
\begin{equation}
\mathrm{RelVar}
=
\frac{\mathrm{Var}(\|\hat{\boldsymbol{g}}\|)}
{\|\boldsymbol{g}\|^2}.
\end{equation}
A smaller value indicates a more stable estimator.

% \begin{figure}[t]
% \centering
% \includegraphics[height=0.49\linewidth]{pictures/relative_variance.pdf}
% \caption{Relative variance of ZO gradient estimates.}
% \label{fig:relative-variance}
% \end{figure}
\begin{figure}[t]
\centering
\includegraphics[width=\linewidth]{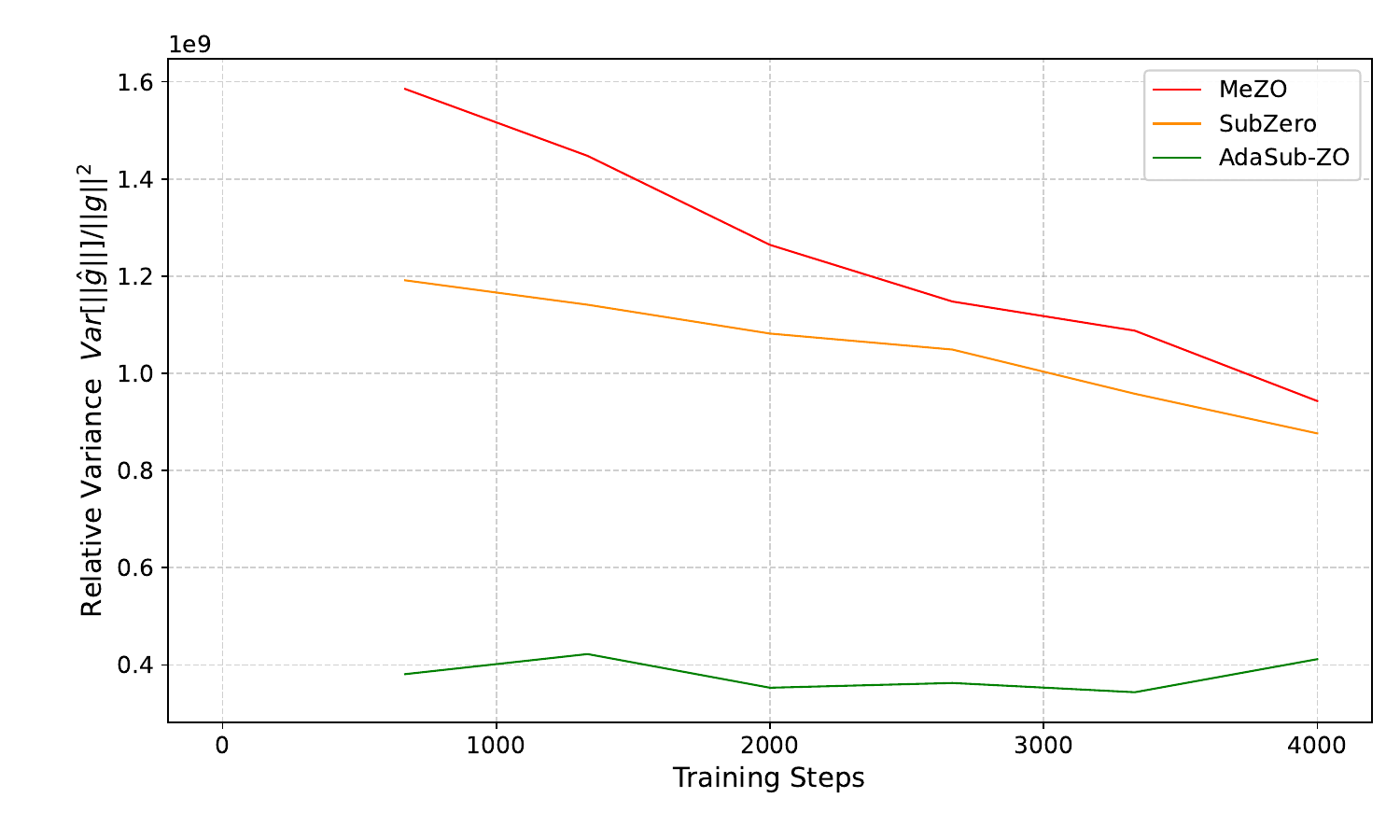}
\caption{Relative variance of ZO gradient estimates.}
\label{fig:relative-variance}
\end{figure}

Figure~\ref{fig:relative-variance} compares the relative variance, and we observe that:
% (1) AdaSub-ZO
% substantially reduces estimation variance.
% Both AdaSub-ZO and its non-momentum variant keep relative variance around
% $0.3\times10^{9}$ to $0.45\times10^{9}$, while MeZO and SubZero are much higher.
% This confirms that the learned low-rank subspace reduces the randomness of ZO
% estimation more effectively than full-dimensional or fixed random perturbations.
% (2) Momentum mainly stabilizes the update, not the sampled estimator.
% The full method and the non-momentum variant have similar single-step variance,
% because both use the same learned low-rank perturbation space. The performance
% gap in Table~\ref{tab:main-results} therefore comes from the momentum-smoothed
% update direction rather than from changing the one-step estimator itself.
(1) AdaSub-ZO substantially reduces the variance of ZO gradient estimates. MeZO shows the largest relative variance due to full-space perturbations, while SubZero remains much higher than AdaSub-ZO, as its low-rank subspaces are randomly sampled. This shows that learning low-rank projection bases can effectively reduce finite-difference noise.
(2) AdaSub-ZO produces more stable update directions. Compared with MeZO and SubZero, AdaSub-ZO keeps the relative variance within a lower and narrower range throughout training. This suggests that the learned subspace better aligns with task-specific descent directions, while subspace momentum further smooths noisy ZO estimates.

\subsection{Training Dynamics}

We compare the training loss curves of MeZO and AdaSub-ZO on the CB task. The curves use moving averages, and the shaded regions
represent training fluctuations.

% \begin{figure}[t]
% \centering
% \includegraphics[height=0.49\linewidth]{pictures/CB_cmp_mezo_adasub_mosub.pdf}
% \caption{Training loss curves on CB.}
% \label{fig:training-loss}
% \end{figure}
\begin{figure}[t]
\centering
\includegraphics[width=\linewidth]{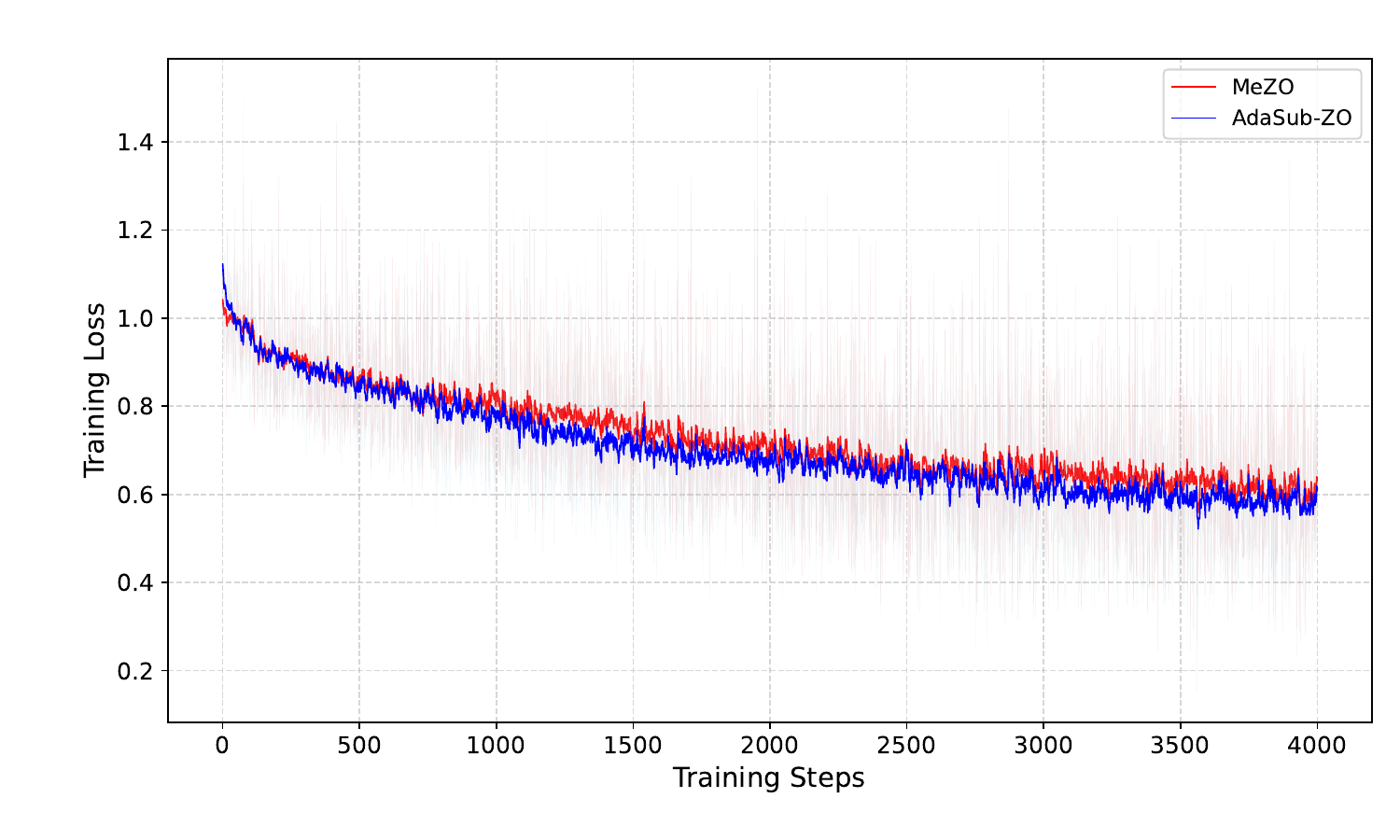}
\caption{Training loss curves on CB.}
\label{fig:training-loss}
\end{figure}

Figure~\ref{fig:training-loss} shows the training dynamics, and we observe that:
% (1) The learned subspace creates a delayed but persistent
% advantage.
% At the beginning of training, MeZO and AdaSub-ZO have similar loss because the
% projection bases are not yet sufficiently adapted. After several subspace
% learning cycles, AdaSub-ZO separates from MeZO and keeps decreasing more
% steadily, indicating that the learned bases gradually capture more useful
% descent directions. (2) Subspace momentum accelerates early and middle-stage
% convergence.
% Full AdaSub-ZO decreases faster than the non-momentum variant after the momentum
% buffer has accumulated enough historical signal. This supports the design of
% maintaining momentum in the $r\times r$ subspace instead of discarding historical
% information after each finite-difference step.
(1) AdaSub-ZO converges faster and reaches a lower training loss on the CB task. Compared with MeZO, AdaSub-ZO reaches a lower training loss after the early stage and maintains this advantage through most of the optimization process. Around 1000 training steps, AdaSub-ZO already obtains a lower loss and converges to a lower final loss at 4000 steps. This indicates that the learned low-rank subspace provides more effective update directions for CB.
(2) AdaSub-ZO shows more stable optimization behavior. MeZO has stronger loss fluctuations and more frequent upward spikes due to full-dimensional random perturbations. In contrast, AdaSub-ZO follows a smoother downward trajectory and keeps the loss in a lower range during later training. This suggests that the learned perturbation subspace and subspace momentum reduce noisy finite-difference updates and stabilize optimization.

\begin{figure}[t]
\centering
\includegraphics[width=0.82\linewidth]{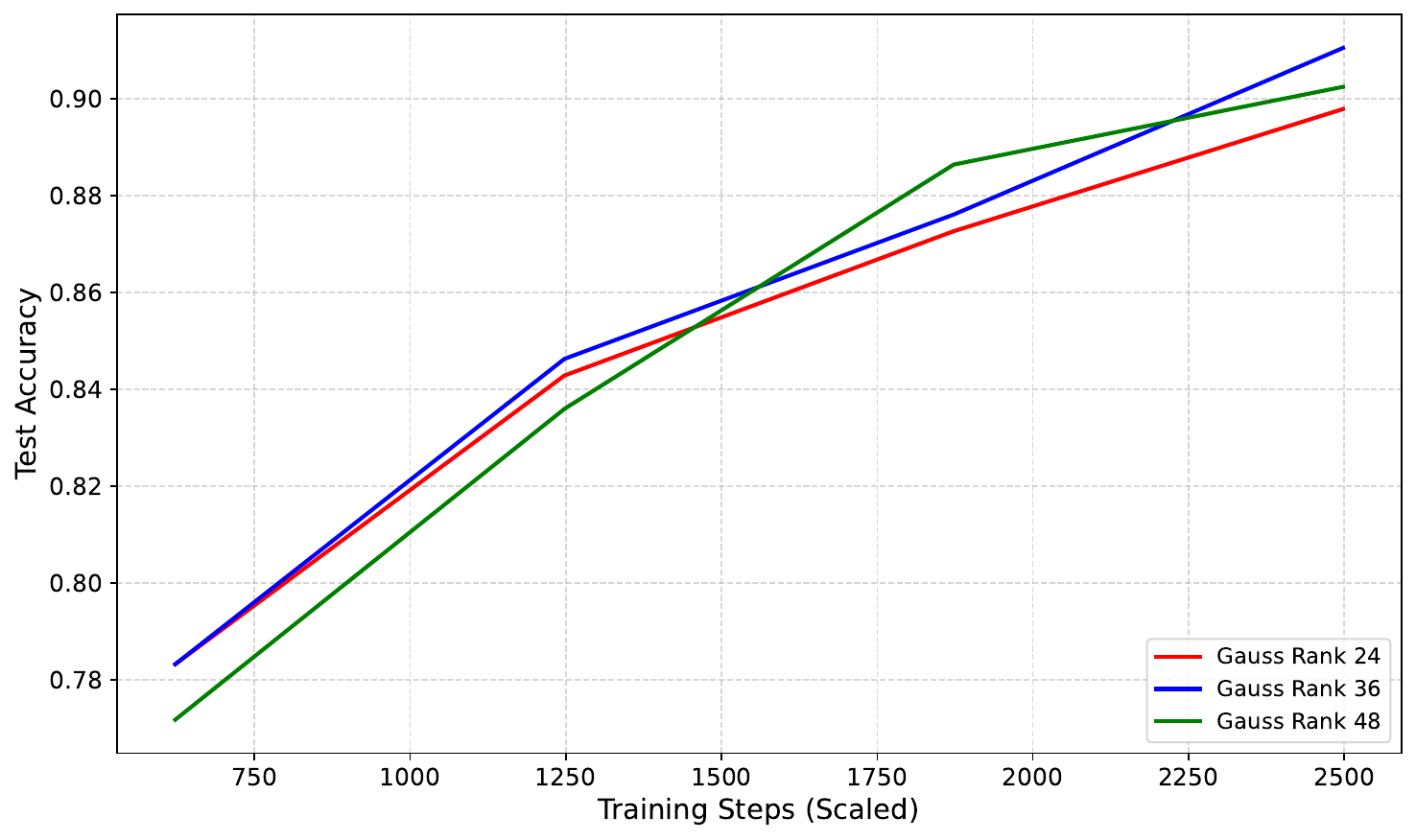}
\caption{Effect of low-rank dimension $r$ on SST-2.}
\label{fig:rank-sensitivity}
\end{figure}

\begin{figure}[t]
\centering
\includegraphics[width=0.82\linewidth]{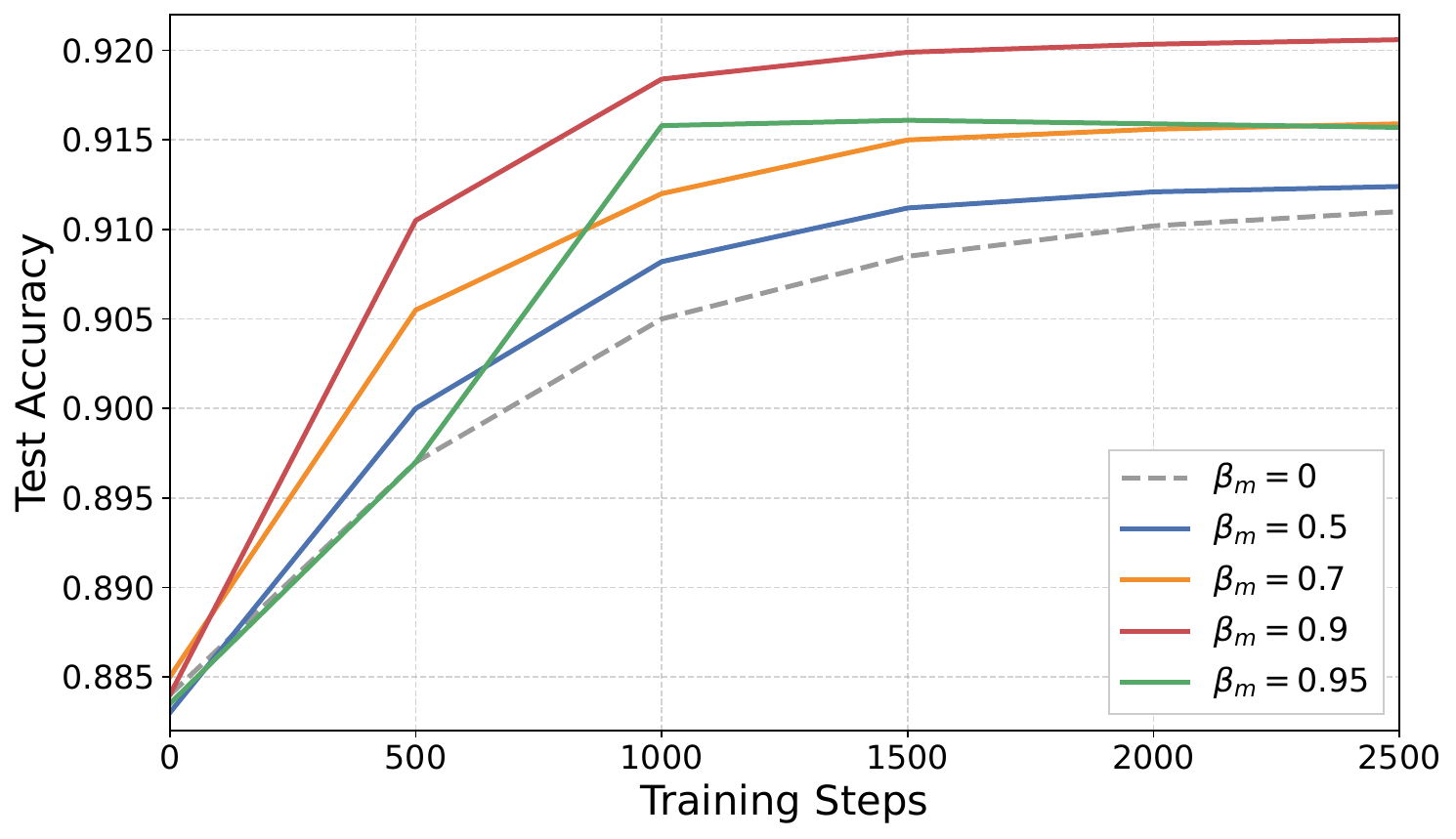}
\caption{Effect of subspace momentum coefficient $\beta_m$ on SST-2.}
\label{fig:momentum-sensitivity}
\end{figure}

\begin{table*}[!t]
\footnotesize
\centering
\caption{Ablation study of learnable module on OPT-1.3B. AVG denotes the average relative improvement over AdaSub-ZO w/o Learnable Module.}
\label{tab:ablation-learnable-module}
\renewcommand{\arraystretch}{1.12}
\setlength{\tabcolsep}{4pt}
\begin{tabular}{l|ccccccc|c}
\toprule
Method & SST-2 & RTE & CB & BoolQ & WSC & WIC & COPA & AVG. \\
\midrule
AdaSub-ZO w/o Learnable Module
& 91.7 & 61.2 & 71.1 & 64.5 & 59.2 & 56.6 & 76.7 & 0.00\% \\
AdaSub-ZO
& \textbf{92.0} & \textbf{61.6} & \textbf{71.4} & \textbf{64.8} & \textbf{59.6} & \textbf{57.0} & \textbf{77.0} & \textbf{+0.52\%} \\
\bottomrule
\end{tabular}
\end{table*}

\begin{table*}[!t]
\footnotesize
\centering
\caption{Ablation study of subspace momentum on OPT-1.3B. AVG denotes the average relative improvement over AdaSub-ZO w/o Subspace Momentum.}
\label{tab:ablation-subspace-momentum}
\renewcommand{\arraystretch}{1.12}
\setlength{\tabcolsep}{4pt}
\begin{tabular}{l|ccccccc|c}
\toprule
Method & SST-2 & RTE & CB & BoolQ & WSC & WIC & COPA & AVG. \\
\midrule
AdaSub-ZO w/o Subspace Momentum
& 91.1 & 61.4 & 70.0 & \textbf{64.9} & \textbf{63.0} & 57.0 & 74.0 & 0.00\% \\
AdaSub-ZO
& \textbf{92.0} & \textbf{61.6} & \textbf{71.4} & 64.8 & 59.6 & 57.0 & \textbf{77.0} & \textbf{+0.26\%} \\
\bottomrule
\end{tabular}
\end{table*}

\subsection{Hyperparameter Sensitivity}

We study two important hyperparameters on SST-2: the low-rank dimension $r$ and the
subspace momentum coefficient $\beta_m$. For $r$, we test
$r\in\{24,36,48\}$. For $\beta_m$, we test
$\beta_m\in\{0,0.5,0.7,0.9,0.95\}$.

Figure~\ref{fig:rank-sensitivity} and Figure~\ref{fig:momentum-sensitivity} show the effects of low-rank dimension $r$ and subspace momentum coefficient $\beta_m$, respectively, and we
observe that:
(1) The best performance is achieved when $r=36$. The reason may be that a
smaller rank restricts the capacity of the perturbation subspace, while a larger
rank increases the difficulty of subspace learning and introduces more estimation
noise. Thus, a moderate rank provides a better balance between subspace
expressiveness and optimization stability.
(2) The best performance is achieved when $\beta_m=0.9$, as a small momentum
coefficient provides limited smoothing, while an excessively large coefficient
adapts slowly to newly learned subspace bases. These results suggest
that subspace momentum should balance noise reduction and adaptability in
learnable subspace ZO fine-tuning.

\subsection{Ablation Study}

\noindent\textbf{Effect of Learnable Subspace Projection.}\par
\noindent To assess the effectiveness of learnable subspace projection, we design
a variant, AdaSub-ZO w/o Learnable Module, which removes forward-only projection
learning and keeps the projection bases fixed. The experimental results are
shown in Table~\ref{tab:ablation-learnable-module}. We observe that full
AdaSub-ZO consistently outperforms this variant across all tasks, improving the
average relative gain from 0.00\% to +0.52\%. This demonstrates that learning
projection bases from finite-difference feedback helps construct more effective
low-rank update subspaces.

\noindent\textbf{Effect of Subspace Momentum.}\par
\noindent To investigate the effectiveness of subspace momentum, we design a
variant, AdaSub-ZO w/o Subspace Momentum, which removes the low-rank momentum
buffer and updates parameters only with the current ZO estimate. The experimental
results are shown in Table~\ref{tab:ablation-subspace-momentum}. We observe that
full AdaSub-ZO achieves a higher average relative gain, improving from 0.00\% to
+0.26\%. This indicates that subspace momentum can smooth noisy finite-difference
updates and further improve the stability of learnable low-rank ZO fine-tuning.

\section{Conclusions}

In this paper, we propose AdaSub-ZO, a zeroth-order fine-tuning framework for
LLMs in learnable low-rank subspaces. Instead of relying on fixed random
perturbation subspaces, AdaSub-ZO learns layer-wise projection bases from
forward-only finite-difference feedback and performs model updates within
low-rank subspaces. To stabilize noisy ZO updates under evolving bases,
AdaSub-ZO further maintains subspace momentum and transfers historical update
information across basis refreshes. This design improves the quality and
stability of update directions. Extensive experiments on SuperGLUE with
OPT-1.3B show that AdaSub-ZO achieves the best overall performance among
representative ZO baselines. Further empirical analyses confirm the
effectiveness of learnable projection bases and subspace momentum for stable and
efficient LLM fine-tuning.

\bibliography{adasub_zo_references}

\appendix
\setcounter{secnumdepth}{1}

\section*{Appendix}

\section{\raggedright Time Complexity Analysis}
\label{sec:appendix_time_complexity}
This appendix provides the time complexity analysis for the unified AdaSub-ZO
framework.
Let $C_{\mathrm{fwd}}$ denote the computational cost of one forward pass,
$d_l=m_l n_l$ be the number of trainable parameters in layer $l$, and
$D=\sum_{l=1}^{L}d_l$. Since AdaSub-ZO is a forward-only method, its dominant
cost comes from finite-difference loss evaluations.

For each model-update step, AdaSub-ZO performs two forward passes to compute the
two-point finite-difference signal. The additional cost associated with constructing the low-rank perturbation $U_l Z_l V_l^\top$ and applying the subspace momentum update can be expressed as follows:
\[
    \mathcal{O}\!\left(\sum_{l=1}^{L} d_l r\right),
\]
while updating the momentum buffers in the $r\times r$ subspace costs only
$\mathcal{O}(Lr^2)$.

Every $K$ iterations, AdaSub-ZO performs $M$ projection-learning steps. Each
projection-learning step also uses two forward passes and incurs
$\mathcal{O}(\sum_{l=1}^{L} d_l r)$ cost for low-rank perturbation construction.
After projection learning, the QR orthogonalization cost of the projection matrices can be expressed as follows:
\[
    \mathcal{O}\!\left(
    \sum_{l=1}^{L}(m_l+n_l)r^2
    \right),
\]
which is lightweight when $r\ll \min(m_l,n_l)$. The momentum coordinate transfer
has the same low-rank order and is also dominated by this QR-related term.

Therefore, over $T$ training iterations, the total time complexity of AdaSub-ZO can be written as follows:
\[
\begin{aligned}
\mathcal{O}\Bigg(
&\left(2T+2M\left\lfloor\frac{T}{K}\right\rfloor\right)C_{\mathrm{fwd}} \\
&+
\left(T+M\left\lfloor\frac{T}{K}\right\rfloor\right)
\sum_{l=1}^{L} d_l r \\
&+
\left\lfloor\frac{T}{K}\right\rfloor
\sum_{l=1}^{L}(m_l+n_l)r^2
\Bigg).
\end{aligned}
\]
Since forward passes dominate the computational cost in LLM fine-tuning and
$r$ is much smaller than the layer dimensions, AdaSub-ZO preserves the
forward-only computational order of standard ZO methods, with only lightweight
low-rank projection learning and subspace momentum overhead.

\section{\raggedright Convergence Analysis of AdaSub-ZO}
\label{sec:appendix_convergence}

This appendix provides the convergence analysis for the unified AdaSub-ZO
framework. In the main paper, AdaSub-ZO is used to denote both the learnable subspace
module and the subspace momentum module. Therefore, the analysis below treats
the momentum version as the full algorithm. The non-momentum variant is recovered by
setting the momentum coefficient to zero.

Let
\[
    f(\theta)=\mathbb{E}_{\mathcal{B}}[\mathcal{L}(\theta;\mathcal{B})],
\]
where $\theta\in\mathbb{R}^{D}$ denotes all trainable parameters. For the
$l$-th layer, AdaSub-ZO constructs
$\widetilde Z_{t,l}=U_{t,l}Z_{t,l}V_{t,l}^{\top}$, where
$U_{t,l}\in\mathbb{R}^{m_l\times r}$,
$V_{t,l}\in\mathbb{R}^{n_l\times r}$, and
$Z_{t,l}\in\mathbb{R}^{r\times r}$ has i.i.d.\ entries sampled from the standard normal distribution. By vectorization,
\[
    \mathrm{vec}(\widetilde Z_{t,l})
    =
    (V_{t,l}\otimes U_{t,l})\mathrm{vec}(Z_{t,l}).
\]
Define $P_{t,l}=V_{t,l}\otimes U_{t,l}\in\mathbb{R}^{d_l\times r^2}$ and
\[
    P_t=\operatorname{bdiag}(P_{t,1},\ldots,P_{t,L})\in\mathbb{R}^{D\times q},
    \qquad q=Lr^2.
\]
The QR step gives $U_{t,l}^{\top}U_{t,l}=I_r$ and
$V_{t,l}^{\top}V_{t,l}=I_r$, so that $P_t^{\top}P_t=I_q$. Thus, the layer-wise
low-rank perturbation in AdaSub-ZO is equivalent to perturbing the parameter
vector along $P_t z_t$, where $z_t\in\mathbb{R}^{q}$ and
$z_t\sim\mathcal{N}(0,I_q)$. The layer-dependent
normalization factors in Eq.~\eqref{eq:adasub_param_update} only rescale the
step size; for clarity, they are absorbed into the effective learning rate in
the analysis.

\noindent\textbf{Assumption 1.}
The objective $f$ is $L_1$-smooth and lower bounded by $f^\star$.

\noindent\textbf{Assumption 2.}
There exists $\kappa\in(0,1]$ such that, at every model-update step,
\[
    \|P_t^{\top}\nabla f(\theta_t)\|^2
    \geq
    \kappa\|\nabla f(\theta_t)\|^2.
\]
The constant $\kappa$ measures how well the learned subspace covers the full
gradient direction. A random isotropic subspace has coverage about $q/D$ in
expectation, while AdaSub-ZO aims to increase this coverage through
forward-only projection learning.

\noindent\textbf{Lemma 1 (Smoothness in a fixed subspace).}
For any fixed $P_t$ satisfying $P_t^{\top}P_t=I_q$, define
$h_t(y)=f(\theta_t+P_t y)$. If $f$ is $L_1$-smooth, then $h_t$ is also
$L_1$-smooth.

\noindent\textit{Proof.}
For any $y_1,y_2\in\mathbb{R}^{q}$,
\[
\begin{aligned}
\Delta_f
&=
\nabla f(\theta_t+P_t y_1)\\
&\quad-
\nabla f(\theta_t+P_t y_2),\\
\|\nabla h_t(y_1)-\nabla h_t(y_2)\|
&=
\|P_t^{\top}\Delta_f\| \\
&\leq
L_1\|P_t(y_1-y_2)\| \\
&=
L_1\|y_1-y_2\|.
\end{aligned}
\]
The last equality uses $P_t^{\top}P_t=I_q$. \hfill $\square$

\noindent\textbf{Lemma 2 (One-step descent without momentum).}
Consider one model-update step with fixed $P_t$ and the two-point estimator:
\[
    \widehat g_t=
    \frac{f(\theta_t+\epsilon P_t z_t)-f(\theta_t-\epsilon P_t z_t)}
    {2\epsilon}P_t z_t.
\]
Let $C_{\epsilon}$ be defined as follows:
\[
    C_{\epsilon}
    =
    \frac{\epsilon^2(q+6)^3}{8}L_1^2
    +
    \frac{3\epsilon^2(q+4)}{32}L_1.
\]
If $\eta_{\theta}=1/[4(q+4)L_1]$, then the following bound holds:
\[
\mathbb{E}\!\left[f(\theta_{t+1})\mid\theta_t,P_t\right]
\leq
f(\theta_t)
-
\frac{\eta_{\theta}}{4}
\|P_t^{\top}\nabla f(\theta_t)\|^2
+
C_{\epsilon}.
\]

\noindent\textit{Proof.}
By Lemma 1, $h_t(y)=f(\theta_t+P_t y)$ is $L_1$-smooth. The original-space
update is equivalent to applying one two-point ZO-SGD step to $h_t$ from
$y=0$. Applying the standard fixed-subspace two-point descent bound
\cite{nesterov2017random,duchi2015optimal}
gives the result: $\nabla h_t(0)=P_t^{\top}\nabla f(\theta_t)$.
\hfill $\square$

\noindent\textbf{Theorem 1 (AdaSub-ZO without subspace momentum).}
Let $\mathcal{S}_T$ be the set of iterations where the model parameters are
updated and let $N_T=|\mathcal{S}_T|$. Under Assumptions 1 and 2, if
$\eta_{\theta}=1/[4(q+4)L_1]$, then the non-momentum AdaSub-ZO variant satisfies:
\[
\begin{aligned}
\frac{1}{N_T}\sum_{t\in\mathcal{S}_T}
\mathbb{E}\|\nabla f(\theta_t)\|^2
\leq&
\frac{16(q+4)L_1(f(\theta_0)-f^\star)}{\kappa N_T} \\
&+
\frac{2\epsilon^2(q+6)^3(q+4)L_1^3}{\kappa}
\\
&+
\frac{3\epsilon^2(q+4)^2L_1^2}{2\kappa}.
\end{aligned}
\]

\noindent\textit{Proof.}
Taking expectation in Lemma 2 and summing over
$t\in\mathcal{S}_T$ yields:
\[
\mathbb{E}[f(\theta_T)]
\leq
f(\theta_0)
-
\frac{\eta_{\theta}}{4}
\sum_{t\in\mathcal{S}_T}
\mathbb{E}\|P_t^{\top}\nabla f(\theta_t)\|^2
+
N_T C_{\epsilon}.
\]
Using $f(\theta_T)\geq f^\star$ and Assumption 2, we obtain:
\[
\frac{\eta_{\theta}\kappa}{4}
\sum_{t\in\mathcal{S}_T}
\mathbb{E}\|\nabla f(\theta_t)\|^2
\leq
f(\theta_0)-f^\star+N_T C_{\epsilon}.
\]
Dividing by $N_T$ and substituting the definitions of $\eta_{\theta}$ and
$C_{\epsilon}$ proves the theorem. \hfill $\square$

Theorem 1 shows that, when $\epsilon$ is sufficiently small, AdaSub-ZO reaches a
first-order stationary point with
$N_T=\mathcal{O}(q/(\kappa\delta))$ update steps. If the subspace is random, then
$\kappa\approx q/D$ and the complexity reduces to the usual
$\mathcal{O}(D/\delta)$ dependence. When projection learning improves subspace
coverage so that $\kappa>q/D$, AdaSub-ZO obtains a better convergence constant
than random-subspace ZO methods.

We next incorporate subspace momentum into the convergence analysis. For the
full AdaSub-ZO algorithm, the subspace-coordinate two-point estimate
$u_t\in\mathbb{R}^{q}$ is defined as follows:
\[
    u_t=
    \frac{f(\theta_t+\epsilon P_t z_t)-f(\theta_t-\epsilon P_t z_t)}
    {2\epsilon}z_t.
\]
AdaSub-ZO maintains a momentum vector $m_t\in\mathbb{R}^{q}$:
\[
    m_t=\beta_m\mathcal{T}_t(m_{t-1})+(1-\beta_m)u_t,
\]
where $\mathcal{T}_t$ is the coordinate-transfer operator induced by
Eq.~\eqref{eq:adasub_momentum_transfer}. Here $m_t$ denotes the normalized
subspace buffer; the unnormalized implementation in
Eq.~\eqref{eq:adasub_momentum} differs only by a constant rescaling that is
absorbed into the effective learning rate. As the old and new bases are
orthonormal, the matrices
$(U_t^{\mathrm{new}})^\top U_t^{\mathrm{old}}$ and
$V_t^{\mathrm{old}}(V_t^{\mathrm{new}})^\top$ have spectral norm at most one.
Therefore:
\[
    \|\mathcal{T}_t(m)\|\leq\gamma\|m\|\leq\|m\|,
\]
which means that momentum transfer does not amplify historical errors. With the
effective learning rate $\widetilde\eta_\theta$, the update can be written as:
\[
    \theta_{t+1}=\theta_t-\widetilde\eta_\theta P_tm_t.
\]

\noindent\textbf{Assumption 3.}
Let $a_t=P_t^{\top}\nabla f(\theta_t)$. There exist
$\rho_m\in[0,1)$ and $\Delta_m\geq0$ such that:
\[
    \mathbb{E}\|m_t-a_t\|^2
    \leq
    \rho_m\mathbb{E}\|a_t\|^2+\Delta_m,
\]
where $\Delta_m$ collects the residual finite-difference error, momentum lag, and
subspace-drift error.

\noindent\textbf{Theorem 2 (AdaSub-ZO with subspace momentum).}
Under Assumptions 1--3, suppose the effective learning rate
$\widetilde\eta_\theta$ satisfies the condition $c_m>0$, where
\[
    c_m=
    \frac{3}{4}
    -
    L_1\widetilde\eta_\theta
    -
    (1+L_1\widetilde\eta_\theta)\rho_m
    >0.
\]
Then full AdaSub-ZO satisfies the following bound:
\[
\frac{1}{N_T}
\sum_{t\in\mathcal{S}_T}
\mathbb{E}\|\nabla f(\theta_t)\|^2
\leq
\frac{f(\theta_0)-f^\star}
{\widetilde\eta_\theta c_m\kappa N_T}
+
\frac{(1+L_1\widetilde\eta_\theta)\Delta_m}
{c_m\kappa}.
\]

\noindent\textit{Proof.}
By $L_1$-smoothness and $P_t^{\top}P_t=I_q$, we have:
\[
\begin{aligned}
f(\theta_{t+1})
&\leq
f(\theta_t)
-
\widetilde\eta_\theta
\langle P_t^{\top}\nabla f(\theta_t),m_t\rangle
+
\frac{L_1\widetilde\eta_\theta^2}{2}\|m_t\|^2.
\end{aligned}
\]
Let $e_t=m_t-a_t$. Since
$\langle a_t,m_t\rangle=\|a_t\|^2+\langle a_t,e_t\rangle$ and
$\langle a_t,e_t\rangle\geq-\|a_t\|^2/4-\|e_t\|^2$, we have:
\[
    \langle a_t,m_t\rangle\geq \frac{3}{4}\|a_t\|^2-\|e_t\|^2.
\]
Also, $\|m_t\|^2=\|a_t+e_t\|^2\leq2\|a_t\|^2+2\|e_t\|^2$. Substituting these
bounds gives:
\[
\begin{aligned}
f(\theta_{t+1})
&\leq
f(\theta_t)
-
\widetilde\eta_\theta
\left(\frac{3}{4}-L_1\widetilde\eta_\theta\right)
\|a_t\|^2
\\
&\quad+
\widetilde\eta_\theta
(1+L_1\widetilde\eta_\theta)
\|e_t\|^2.
\end{aligned}
\]
Taking expectation and using Assumption 3, we obtain:
\[
\mathbb{E}f(\theta_{t+1})
\leq
\mathbb{E}f(\theta_t)
-
\widetilde\eta_\theta c_m\mathbb{E}\|a_t\|^2
+
\widetilde\eta_\theta(1+L_1\widetilde\eta_\theta)\Delta_m.
\]
Summing over $\mathcal{S}_T$, using $f(\theta_T)\geq f^\star$, and applying
Assumption 2 complete the proof. \hfill $\square$

Theorem 2 states that the full AdaSub-ZO algorithm converges to a neighborhood
of a first-order stationary point. The first term decays as
$\mathcal{O}(1/(\kappa N_T))$, while the second term is controlled by the
momentum tracking error. For independent finite-difference noise, EMA-style
momentum reduces the noise scale by the factor $(1-\beta_m)/(1+\beta_m)$, so that:
\[
\Delta_m
=
\mathcal{O}\!\left(
\frac{1-\beta_m}{1+\beta_m}\epsilon^2L_1^2q^3
+
\Delta_{\mathrm{drift}}
\right).
\]
Thus subspace momentum smooths noisy ZO directions without maintaining
full-dimensional optimizer states, and the non-expansive transfer keeps
historical momentum stable across evolving learned bases.

\section{\raggedright Hyperparameter Settings}
\label{sec:appendix_hyperparams}

This appendix provides the hyperparameter settings used in our experiments.
To ensure a fair comparison and optimal performance, we performed a grid search on the validation set to determine the hyperparameters for AdaSub-ZO. The hyperparameter search space for our unified AdaSub-ZO framework, incorporating both the learnable low-rank projection and subspace momentum modules, is summarized in Table~\ref{tab:hyper-adasub-unified}.

\begin{table}[H]
\centering
\caption{Hyperparameter search space for AdaSub-ZO.}
\label{tab:hyper-adasub-unified}
\renewcommand{\arraystretch}{1.15}
\resizebox{\columnwidth}{!}{
\begin{tabular}{@{}l|l@{}}
\toprule
\textbf{Hyperparameter} & \textbf{Search Space} \\ \midrule
Model Learning Rate $\eta_\theta$ & $\{1\text{e-}7, 1\text{e-}6, 1\text{e-}5\}$ \\
Model Perturbation Scale $\epsilon$ & $\{1\text{e-}4, 1\text{e-}3, 1\text{e-}2\}$ \\
Low-Rank Dimension $r$ & $\{24, 36, 48\}$ \\
Projection Learning Cycle $K$ & $\{50, 100, 200, 500\}$ \\
Projection Learning Steps $M$ & $\{1, 5, 10, 20\}$ \\
Projection Learning Rate $\eta_\phi$ & $\{1\text{e-}3, 1\text{e-}2, 1\text{e-}1\}$ \\
Projection Perturbation Scale $\mu$ & $\{1\text{e-}3, 1\text{e-}2, 1\text{e-}1\}$ \\
Subspace Momentum Coefficient $\beta_m$ & $\{0.5, 0.7, 0.9, 0.95\}$ \\
Projection EMA Coefficient $\beta$ & $\{0.5, 0.7, 0.9\}$ \\
\bottomrule
\end{tabular}
}
\end{table}

For the baseline methods (MeZO, ZO-AdaMU, LOZO, and SubZero), we strictly followed the official implementations and their recommended hyperparameter search spaces. Specifically, to ensure that the performance gains strictly come from our proposed modules,
AdaSub-ZO uses the same learning-rate search space $\eta_\theta \in \{1\text{e-}7, 1\text{e-}6, 1\text{e-}5\}$ and
finite-difference perturbation-scale search space $\epsilon \in \{1\text{e-}4, 1\text{e-}3, 1\text{e-}2\}$ as MeZO and SubZero. The low-rank dimension $r$ is aligned with SubZero's settings to ensure the perturbation subspace dimensions remain identical.

\end{document}